\pdfoutput=1
\documentclass{article}
\usepackage{iclr2026_conference,times}

\usepackage{amsmath,mathtools,amssymb,amsfonts,amsthm}
\usepackage{graphicx}
\usepackage{subcaption}
\usepackage{booktabs}
\usepackage{microtype}
\usepackage{enumitem}
\usepackage{listings}
\usepackage{xcolor}
\usepackage{xfrac}
\usepackage{hyperref}
\hypersetup{
  pdftitle={A Fix for Softmax and Cross-Entropy},
  pdfauthor={Jonathan T. Barron, Dan Gnanapragasam, Michael Tschannen}
}
\usepackage{url}

\setlist[itemize]{wide=0pt,itemsep=1.5pt,topsep=2.5pt}
\setlist[enumerate]{wide=0pt,itemsep=1.5pt,topsep=2.5pt}
\graphicspath{{./}{../figures/out/}}

\newcommand\lft{\mathopen{}\left}
\newcommand\rgt{\aftergroup\mathclose\aftergroup{\aftergroup}\right}
\makeatletter
\let\origsum\sum
\renewcommand{\sum}{\DOTSB\mathop{\mathchoice{\origsum}{\raisebox{0pt}[7.0pt][1.0pt]{$\textstyle\origsum$}}{\origsum}{\origsum}}\slimits@}
\makeatother
\newcommand{\bfsixteen}{bf16}
\newcommand{\fpthirtytwo}{fp32}
\newcommand{\fpsixtyfour}{fp64}
\newcommand{\softmaxtxt}{softmax}
\newcommand{\logsoftmaxtxt}{log-softmax}
\newcommand{\crossentropytxt}{cross-entropy}
\DeclareMathOperator*{\LSE}{LSE}
\newcommand{\logonep}{\operatorname{log1p}}
\newcommand{\expmone}{\operatorname{expm1}}
\newcommand{\softplus}{\operatorname{softplus}}
\newcommand{\softmax}{\operatorname{softmax}}
\newcommand{\logsoftmax}{\operatorname{logsoftmax}}
\newcommand{\numclasses}{K}
\newcommand{\targetidx}{t}
\newcommand{\imax}{i_{\max}}
\newcommand{\cliff}[1]{c_{#1}}
\newcommand{\cliffd}{\cliff{d}}
\newcommand{\cliffsixteen}{\cliff{16}}
\newcommand{\cliffthirtytwo}{\cliff{32}}
\newcommand{\cliffsixtyfour}{\cliff{64}}
\newcommand{\vlogits}{\mathbf{z}}
\newcommand{\logits}{z}
\newcommand{\shiftedlogits}{\tilde{z}}
\newcommand{\zmax}{z_{\max}}
\newcommand{\vpredprob}{\mathbf{q}}
\newcommand{\predprob}{q}
\newcommand{\vtargetprob}{\mathbf{p}}
\newcommand{\targetprob}{p}
\newcommand{\vzero}{\mathbf{0}}
\newcommand{\loss}{\mathcal{L}}
\newcommand{\vgrad}{\mathbf{g}}
\newcommand{\grad}{g}
\newcommand{\vfgrad}{\tilde{\mathbf{g}}}
\newcommand{\fgrad}{\tilde{g}}
\newcommand{\margin}{m}
\newcommand{\eqmargin}{m^\star}
\newcommand{\zstudent}{\mathbf{z}_{\mathrm{S}}}
\newcommand{\zteacher}{\mathbf{z}_{\mathrm{T}}}
\newcommand{\zstudentk}[1]{z_{\mathrm{S},#1}}
\newcommand{\qstudent}{\mathbf{q}_{\mathrm{S}}}
\newcommand{\pteacher}{\mathbf{p}_{\mathrm{T}}}
\newcommand{\paramsstudent}{\boldsymbol{\theta}_{\mathrm{S}}}
\newcommand{\paramsteacher}{\boldsymbol{\theta}_{\mathrm{T}}}
\newcommand{\temp}{T}
\newcommand{\vlogitdiff}{\boldsymbol{\delta}}
\newcommand{\logitdiff}{\delta}
\newcommand{\centereddiff}{d}
\newcommand{\logratio}{s}
\newcommand{\probdiff}{u}
\newcommand{\vparams}{\boldsymbol{\theta}}
\newcommand{\params}{\theta}
\newcommand{\vadamfirst}{\hat{\mathbf{m}}}
\newcommand{\adamfirst}{\hat{m}}
\newcommand{\vadamsecond}{\hat{\mathbf{v}}}
\newcommand{\adamsecond}{\hat{v}}
\newcommand{\adameps}{\epsilon}
\newcommand{\wdparam}{\lambda}
\newcommand{\lrparam}{\eta}
\newcommand{\betaone}{\beta_1}
\newcommand{\betatwo}{\beta_2}

\title{A Fix for Softmax and Cross-Entropy}

\author{Jonathan T. Barron \\
Google DeepMind \\
\texttt{barron@google.com}
\And
Dan Gnanapragasam \\
Google DeepMind \\
\texttt{nanabyte@google.com}
\And
Michael Tschannen \\
Google DeepMind \\
\texttt{tschannen@google.com}
}

\iclrfinalcopy

\begin{document}

\maketitle
\lhead{}

\begin{abstract}
Standard implementations of \softmaxtxt{}, \logsoftmaxtxt{}, and
\crossentropytxt{} have two numerical bugs.
The first bug, \emph{Softmax Collapse}, rounds the target logit's gradient to
zero if that one logit is much larger than the others, resulting in
gradients that no longer sum to zero.
The second bug, \emph{Softmax Cancellation},
occurs when student logits are close to teacher logits,
and produces non-zero gradients even when the student and teacher logits are
equal.
We present simple implementations of \logsoftmaxtxt{}, \softmaxtxt{}, and
\crossentropytxt{} that fix both bugs.
Our fix eliminates the previously observed ``slingshot'' phenomenon,
speeds up grokking with weight decay, prevents accuracy collapse without weight
decay, stops student networks from drifting away from the teacher during
distillation, and shows that after memorization, logit growth is bounded by an
equilibrium between weight decay and Adam's $\adameps$ hyperparameter.
\end{abstract}

\section{Introduction}
\label{sec:intro}

\begin{figure}[b!]
    \centering
    \includegraphics[width=0.68\textwidth]{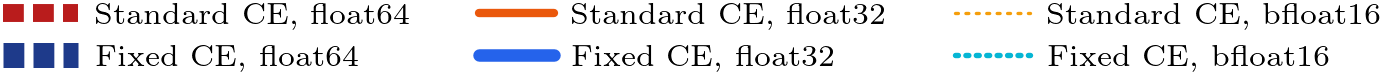}\\[2pt]
    \begin{subfigure}[t]{0.4809\textwidth}
        \centering
        \includegraphics[width=\textwidth]{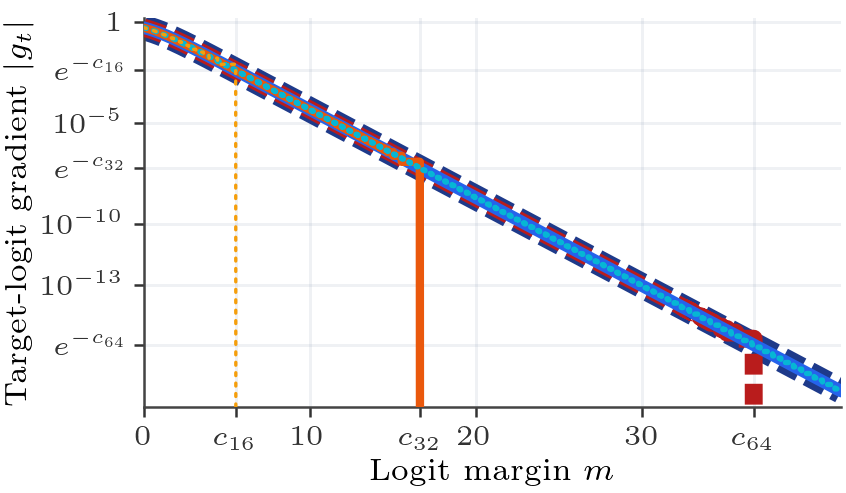}
        \captionsetup{margin={0.0794\textwidth,0pt}}
        \caption{%
          Softmax Collapse: $|\grad_\targetidx|$ vs.\ margin $\margin$
        }
        \label{fig:fig1a}
    \end{subfigure}\hfill
    \begin{subfigure}[t]{0.4911\textwidth}
        \centering
        \includegraphics[width=\textwidth]{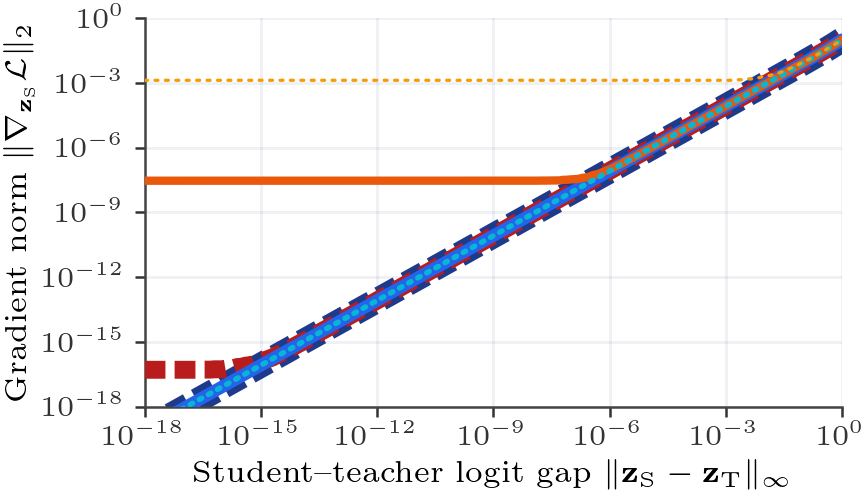}
        \captionsetup{margin={0.0703\textwidth,0pt}}
        \caption{%
          Softmax Cancellation: $\|\nabla_{\zstudent}\loss\|_2$ as $\zstudent
          \to \zteacher$
        }
        \label{fig:fig4a}
    \end{subfigure}
    \vspace{-4pt}
    \caption{%
      The two bugs in standard \softmaxtxt{} and \crossentropytxt{}.
      \textbf{(a) Softmax Collapse:} As a model becomes more confident in the
      correct class (thereby increasing the logit margin $\margin$), the true
      gradient on the target logit shrinks exponentially, but standard
      implementations incorrectly round that gradient to zero once $\margin$
      crosses a precision cliff ($\approx 5.5$ in \bfsixteen{}, $16.6$ in
      \fpthirtytwo{}, and $36.7$ in \fpsixtyfour{}).
      \textbf{(b) Softmax Cancellation:} In soft-target distillation, as
      student logits $\zstudent$ approach teacher logits $\zteacher$, the true
      gradient shrinks toward zero, but standard implementations lose precision
      and plateau at a non-zero noise floor.
      Our fix eliminates both bugs and yields accurate gradients across all
      precisions.
    }
    \label{fig:fig1}
    \vspace{-4pt}
\end{figure}

Naively computing \softmaxtxt{} or \crossentropytxt{} on logits $\vlogits \in
\mathbb{R}^\numclasses$ as $e^{\logits_k} / \sum_j e^{\logits_j}$ overflows if
any logit exceeds the floating-point exponent limit ($\approx 88$ in
\fpthirtytwo{}), which is why standard implementations in all deep learning
frameworks (JAX, PyTorch, TensorFlow, etc.) shift the input logits by
subtracting their maximum value $\zmax = \max_j \logits_j$
\citep{higham2002accuracy,blanchard2021accurately}:
\begin{equation}
    \shiftedlogits_k = \logits_k - \zmax\,, \quad
    \logsoftmax(\vlogits)_k
    = \shiftedlogits_k - \log \sum_j e^{\shiftedlogits_j}\,,
    \quad
    \predprob_k = \softmax(\vlogits)_k
    = \frac{e^{\shiftedlogits_k}}
           {\sum_j e^{\shiftedlogits_j}}\,.
    \label{eq:std_softmax}
\end{equation}
For a target distribution $\vtargetprob \in \Delta^{\numclasses-1}$,
\crossentropytxt{} (CE) is $\loss(\vlogits, \vtargetprob) =
-\sum_k \targetprob_k\,\logsoftmax(\vlogits)_k$, with exact gradient
$\vgrad = \nabla_{\vlogits} \loss(\vlogits, \vtargetprob) = \vpredprob -
\vtargetprob$.
Though subtracting $\zmax$ prevents overflow, two numerical bugs remain.

The first bug, identified by \citet{prieto2025grokking} as \emph{Softmax
Collapse} (Figure~\ref{fig:fig1a}), occurs when a
model correctly predicts the target class $\targetidx \in \{1, \dots,
\numclasses\}$ with high confidence, so that $\logits_\targetidx = \zmax$ is
much larger than all non-target logits $\logits_j$ ($j \neq \targetidx$).
We measure this confidence by the \emph{logit margin}, which is how much larger
the target logit is than the log-sum-exp of all non-target logits:
\begin{equation}
    \margin = \logits_\targetidx
    - \log \sum_{j \neq \targetidx} e^{\logits_j}\,.
    \label{eq:margin}
\end{equation}
In Equation~\ref{eq:std_softmax}, subtracting $\zmax = \logits_\targetidx$ sets
the shifted target logit to $\shiftedlogits_\targetidx = 0$ (so
$e^{\shiftedlogits_\targetidx} = 1$ in the denominator), while the $\numclasses
- 1$ non-target exponentials sum to a small positive number $e^{-\margin} =
\sum_{j \neq \targetidx} e^{\logits_j - \logits_\targetidx} \ll 1$.
Floating-point numbers have much finer precision near $0$ than near $1$: in
\fpthirtytwo{}, values near $0$ can be as small as ${\sim}10^{-38}$, whereas
values added to $1.0$ are lost to rounding below ${\sim}10^{-7}$.
Because standard implementations form the denominator $1 + e^{-\margin}$ by
adding $e^{-\margin}$ to $1.0$, the $e^{-\margin}$ term is lost to rounding
once the margin $\margin$ exceeds a threshold $\cliffd$ that we will call the
\emph{precision cliff} ($\cliffsixteen \approx 5.5$ in \bfsixteen{},
$\cliffthirtytwo \approx 16.6$ in \fpthirtytwo{}, and $\cliffsixtyfour \approx
36.7$ in \fpsixtyfour{}).
When $\margin > \cliffd$, the denominator $1 + e^{-\margin}$ evaluates to
exactly $1.0$, which causes $\predprob_\targetidx = 1/(1 + e^{-\margin})$ to
evaluate to $1.0$ and zeroes out both the loss and the target gradient
$\fgrad_\targetidx = \predprob_\targetidx - 1 = 0.0$.
However, the non-target gradients $\fgrad_j = \predprob_j = e^{\logits_j -
\logits_\targetidx} / (1 + e^{-\margin})$ ($j \neq \targetidx$) are evaluated
near $0$ without adding $1.0$ to their numerators, so they remain positive down
to exponent underflow ($\margin \approx 88$).
Because the target gradient rounds to $0.0$ while the non-target gradients do
not, the gradients no longer sum to zero ($\sum_k \fgrad_k > 0$).
When training with adaptive optimizers such as Adam and AdamW
\citep{kingma2015adam,loshchilov2019decoupled}, which normalize each gradient
by a moving average of its magnitude, this asymmetry causes the recurring
``slingshot'' loss spikes \citep{thilak2022slingshot,liu2026grokking} and
stalls grokking \citep{prieto2025grokking}.
While prior works used surrogate losses \citep{prieto2025grokking}, gradient
projection \citep{liu2026grokking}, or \fpsixtyfour{} casting
\citep{nanda2023progress,varma2023explaining} to ameliorate these issues, we
instead fix \logsoftmaxtxt{}, \softmaxtxt{}, and \crossentropytxt{} directly
(Section~\ref{sec:bug1}).

The second bug, which we will call \emph{Softmax Cancellation}
(Figure~\ref{fig:fig4a}), occurs in
knowledge distillation \citep{hinton2015distilling} and KL regularization,
where the target $\vtargetprob = \softmax(\zteacher)$ comes from teacher logits
$\zteacher$.
Standard implementations compute $\vpredprob = \softmax(\zstudent)$ and
$\vtargetprob = \softmax(\zteacher)$ separately and subtract them to get
$\vfgrad = \vpredprob - \vtargetprob$.
When $\zstudent$ is close to $\zteacher$, subtracting these
nearly equal probabilities loses precision, and slight differences between the
forward-pass \softmaxtxt{} and backward-pass \logsoftmaxtxt{} formulas leave a
non-zero gradient even when $\zstudent = \zteacher$.
We fix this by computing $\vpredprob - \vtargetprob$ directly from $\zstudent -
\zteacher$ in log space (Section~\ref{sec:bug2}).

\section{Softmax Collapse}
\label{sec:bug1}

When the target logit $\logits_\targetidx = \zmax$ is the largest logit,
shifting by $\zmax$ in Equation~\ref{eq:std_softmax} sets
$\shiftedlogits_\targetidx = 0$ and $\shiftedlogits_j = \logits_j -
\logits_\targetidx < 0$ for $j \neq \targetidx$, so the denominator is
$\sum_i e^{\shiftedlogits_i} = 1 + \sum_{j \neq \targetidx} e^{\logits_j -
\logits_\targetidx} = 1 + e^{-\margin}$.
Mathematically, the hard-target loss is $\loss(\vlogits, \targetidx) =
-\logsoftmax(\vlogits)_\targetidx = \log(1 + e^{-\margin})$, with gradients
\begin{equation}
    \grad_\targetidx
    = \underbrace{\frac{1}{1 + e^{-\margin}} - 1}_{\text{Standard}}
    = \underbrace{-\frac{e^{-\margin}}{1 + e^{-\margin}}}_{\text{Fixed}}\,,
    \qquad
    \grad_j = \frac{e^{\logits_j - \logits_\targetidx}}{1 + e^{-\margin}}
    \quad \text{for } j \neq \targetidx\,,
    \label{eq:hard_exact}
\end{equation}
which sum to zero ($\sum_k \grad_k = 0$).
In floating point, however, the two expressions for $\grad_\targetidx$ in
Equation~\ref{eq:hard_exact} behave very differently once $\margin > \cliffd$,
where $e^{-\margin}$ is too small to change $1.0$ when added to it
(Figure~\ref{fig:fig1a}).
Standard implementations form the denominator $1 + e^{-\margin}$,
which rounds to $1$ when $\margin > \cliffd$
\citep{prieto2025grokking}.
In the forward pass, $\log(1 + e^{-\margin})$ evaluates to $\log(1)
= 0$, so both $\logsoftmax(\vlogits)_\targetidx$ and $\loss(\vlogits,
\targetidx)$ round to zero.
In the backward pass, standard implementations compute $\predprob_\targetidx
= 1/(1 + e^{-\margin}) = 1$ and subtract $1$ to get
$\fgrad_\targetidx = \predprob_\targetidx - 1 = 0$.
In contrast, our fixed expression $-e^{-\margin}/(1 + e^{-\margin})$ keeps
$e^{-\margin}$ in the numerator: even when the denominator $1 + e^{-\margin}$
rounds to $1.0$, dividing $-e^{-\margin}$ by $1.0$ preserves $\grad_\targetidx
\approx -e^{-\margin}$ down to exponent underflow ($\margin \approx 88$).

While $\fgrad_\targetidx$ rounds to zero when $\margin > \cliffd$, the
non-target gradients do not.
For each non-target class $j \neq \targetidx$, standard cross-entropy computes
$\fgrad_j = \predprob_j = e^{\logits_j - \logits_\targetidx} / (1 +
e^{-\margin})$.
Because the numerator $e^{\logits_j - \logits_\targetidx}$ is evaluated directly
from $\logits_j - \logits_\targetidx < 0$ without adding $1$, it remains
accurate down to exponent underflow while the denominator rounds to $1$.
When $\margin > \cliffd$, standard cross-entropy zeroes out only the target
entry $\targetidx$ while all $\numclasses - 1$ non-target gradients remain
positive \citep{liu2026grokking}:
\begin{equation}
    \fgrad_k = \mathbf{1}{[k \neq \targetidx]}\,\grad_k\,, \qquad
    \sum_k \fgrad_k = -\grad_\targetidx
    = \frac{e^{-\margin}}{1 + e^{-\margin}} > 0
    \quad \text{for } \margin > \cliffd\,.
    \label{eq:hard_buggy}
\end{equation}
The same bug affects standalone $\softmax(\vlogits)$ \citep{qiu2026why}: once
$\margin > \cliffd$, its backward pass rounds $\predprob_{\imax}$ ($\imax =
\arg\max_i \logits_i$) to $1$ and zeroes $\fgrad_{\imax}$ while the other
gradients stay non-zero.

This bug can be fixed by never adding $1$ to $e^{-\margin}$ directly.
In $\logsoftmax(\vlogits)$ and $\softmax(\vlogits)$, only the largest logit
$\logits_{\imax}$ is shifted to $0$ (where $e^0 = 1$ in the denominator).
Instead of including $e^0 = 1$ in the sum of exponentials, we set
$\shiftedlogits_{\imax} = 0$, compute the margin $\margin$ by excluding $\imax$
from the log-sum-exp \citep{blanchard2021accurately}, and combine them with a
numerically stable $\softplus(x) = \log(1 + e^x)$ (with
$\softmax(\vlogits)_k = \exp(\logsoftmax(\vlogits)_k)$):
\begin{equation}
    \shiftedlogits_k
    = \mathbf{1}{[k \neq \imax]}{(\logits_k - \logits_{\imax})}\,,
    \;\;
    \margin = -\LSE_{j \neq \imax} \shiftedlogits_j\,,
    \;\;
    \logsoftmax(\vlogits)_k = \shiftedlogits_k - \softplus(-\margin)\,.
    \label{eq:logsoftmax_fix}
\end{equation}
For hard-target \crossentropytxt{} with target $\targetidx$, we can use
$\targetidx$ directly in place of $\imax$ without an $\arg\max$:
\begin{equation}
    \margin = -\LSE_{j \neq \targetidx}(\logits_j - \logits_\targetidx)\,,
    \qquad
    \loss(\vlogits, \targetidx)
    = \log(1 + e^{-\margin})
    = \softplus(-\margin)\,.
    \label{eq:hard_fix}
\end{equation}
Because the $\targetidx$-th slot is replaced with the constant $0$ rather than
$\logits_\targetidx - \logits_\targetidx$, $\logits_\targetidx$ only appears in
$\loss$ through the $\numclasses - 1$ differences $\logits_j -
\logits_\targetidx$ ($j \neq \targetidx$), so autodiff correctly computes the
target gradient as the negated sum of the non-target gradients:
\begin{equation}
    \grad_j = e^{\logits_j - \logits_\targetidx - \loss}
    = \frac{e^{\logits_j - \logits_\targetidx}}{1 + e^{-\margin}}
    \quad \text{for } j \neq \targetidx\,, \qquad
    \grad_\targetidx = -\sum_{j \neq \targetidx} \grad_j
    = -\frac{e^{-\margin}}{1 + e^{-\margin}}\,.
    \label{eq:hard_fix_grad}
\end{equation}
Here is a JAX implementation of \logsoftmaxtxt{}, \softmaxtxt{}, and hard-target
\crossentropytxt{}, which evaluates $\log(1 + e^{-\margin})$ via $\logonep$,
remains accurate down to exponent underflow, and satisfies $\sum_k \grad_k = 0$:

\begin{lstlisting}
import jax, jax.numpy as jnp, optax

def log_softmax(z):
  i_max = jnp.argmax(z, axis=-1)
  z_max = jnp.take_along_axis(z, i_max[..., None], axis=-1)  # Max logit.
  is_max = jax.nn.one_hot(i_max, z.shape[-1], dtype=bool)    # Max mask.
  shifted = jnp.where(is_max, 0.0, z - z_max)                # Exact 0 at argmax.
  exp_others = jnp.where(is_max, 0.0, jnp.exp(shifted))      # Non-max exp.
  s = jnp.sum(exp_others, axis=-1, keepdims=True)            # Sum non-max exp.
  return shifted - jnp.log1p(s)                              # log_softmax(z).

softmax = lambda z: jnp.exp(log_softmax(z))

def cross_entropy(z, t):
  z_t = jnp.take_along_axis(z, t[..., None], axis=-1)        # Target logit.
  is_t = jax.nn.one_hot(t, z.shape[-1], dtype=bool)          # Target mask.
  shifted = jnp.where(is_t, 0.0, z - z_t)                    # Exact 0 at target.
  b = jnp.max(shifted, axis=-1, keepdims=True)               # Max shift >= 0.
  s = jnp.sum(jnp.where(is_t, 0.0, jnp.exp(shifted - b)), axis=-1)
  return b[..., 0] + jnp.log1p(s + jnp.expm1(-b[..., 0]))    # -log_softmax(z)_t.
\end{lstlisting}

As an example, consider $\vlogits = [9.0, -9.0, -9.0]$ and $\targetidx = 0$.
The standard \texttt{jax.nn} and \texttt{optax} implementations return zero for
the target log-probability, loss, and target gradient, while our functions
return accurate values whose gradients sum to zero:

\begin{lstlisting}
z, t = jnp.array([9.0, -9.0, -9.0], dtype=jnp.float32), jnp.array(0)
std_ce = optax.softmax_cross_entropy_with_integer_labels
jax.nn.log_softmax(z)                       # ->           [ 0.0,      -18.0,     -18.0    ]
jax.grad(lambda z: jax.nn.softmax(z)[0])(z) # ->           [ 0.0,      -1.523e-8, -1.523e-8]
jax.value_and_grad(std_ce)(z, t)            # -> 0.0,      [ 0.0,       1.523e-8,  1.523e-8]
log_softmax(z)                              # ->           [-3.046e-8, -18.0,     -18.0    ]
jax.grad(lambda z: softmax(z)[0])(z)        # ->           [ 3.046e-8, -1.523e-8, -1.523e-8]
jax.value_and_grad(cross_entropy)(z, t)     # -> 3.046e-8, [-3.046e-8,  1.523e-8,  1.523e-8]
\end{lstlisting}

\section{Softmax Cancellation}
\label{sec:bug2}

Given student logits $\zstudent \in \mathbb{R}^\numclasses$ and teacher logits
$\zteacher \in \mathbb{R}^\numclasses$, standard soft-target \crossentropytxt{}
computes $\vtargetprob = \softmax(\zteacher)$ and evaluates
\begin{equation}
    \loss(\zstudent, \zteacher)
    = -\sum_k \targetprob_k\,\logsoftmax(\zstudent)_k\,, \quad
    \vpredprob = \softmax(\zstudent)\,, \quad
    \fgrad_k = \frac{\partial \loss}{\partial \zstudentk{k}}
    = \predprob_k - \targetprob_k\,.
    \label{eq:soft_buggy}
\end{equation}
As the student logits approach the teacher logits ($\zstudent \to \zteacher$),
the true gradient $\vgrad = \vpredprob - \vtargetprob$ shrinks toward $\vzero$
and equals $\vzero$ when $\zstudent = \zteacher$.
In floating point, however, computing $\vpredprob$ and $\vtargetprob$ separately
and subtracting them (Equation~\ref{eq:soft_buggy}) loses precision when
$\zstudent$ is close to $\zteacher$, because the small logit difference
$\vlogitdiff = \zstudent - \zteacher$ is lost when $\vpredprob$ and
$\vtargetprob$ are rounded to probabilities before being subtracted
(Figure~\ref{fig:fig4a}).
Furthermore, because $\vtargetprob = \softmax(\zteacher)$ in the forward pass
and $\vpredprob = \exp(\logsoftmax(\zstudent))$ in the backward pass are
evaluated using slightly different sequences of operations, $\vpredprob$ and
$\vtargetprob$ do not match to machine precision even when $\zstudent =
\zteacher$, leaving a non-zero gradient $\vfgrad = \vpredprob - \vtargetprob
\neq \vzero$.

To avoid subtracting nearly equal probabilities, we pass $\zteacher$ directly
to our loss function and compute $\vpredprob - \vtargetprob$ in log space using
$\expmone(x) = e^x - 1$ and $\logonep(x) = \log(1 + x)$.
Given $\vlogitdiff = \zstudent - \zteacher$, we center $\vlogitdiff$ under
$\vtargetprob$ and write the log-ratio $\logratio_k = \log(\predprob_k /
\targetprob_k)$ and difference $\probdiff_k = \predprob_k - \targetprob_k$ as:
\begin{equation}
    \centereddiff_k = \logitdiff_k - \sum_i \targetprob_i \logitdiff_i\,,
    \qquad
    \logratio_k = \centereddiff_k
    - \logonep\lft(\sum_i \targetprob_i \expmone(\centereddiff_i)\rgt)\,,
    \qquad
    \probdiff_k = \targetprob_k \expmone(\logratio_k)\,.
    \label{eq:soft_fix}
\end{equation}
Because $\sum_i \targetprob_i \centereddiff_i = 0$ and $\expmone(x) \ge x$, the
input to $\logonep$ is always non-negative.
When $|\logratio_k| \ge 1$, we fall back to $\probdiff_k = \predprob_k -
\targetprob_k$ to avoid $0.0 \times \infty$ if $\targetprob_k$ rounds to $0.0$
while $\expmone$ overflows, and we set $\grad_k = \probdiff_k - \predprob_k
\sum_i \probdiff_i$ so that $\sum_k \grad_k = 0$.
Because Equation~\ref{eq:soft_fix} avoids adding or subtracting $1$, $\vgrad$
remains accurate as $\zstudent \to \zteacher$ and is $\vzero$ when $\zstudent =
\zteacher$.
Here is a JAX implementation of soft-target \crossentropytxt{} (using
\texttt{log\_softmax} and \texttt{softmax} from Section~\ref{sec:bug1}), which
uses \texttt{jax.lax.stop\_gradient} (\texttt{sg}) to return the forward loss
while substituting $\vgrad$ in the backward pass:

\begin{lstlisting}
def soft_target_cross_entropy(z_S, z_T):
  log_q = log_softmax(z_S)
  q, p = jnp.exp(log_q), softmax(z_T)
  delta = z_S - z_T                                                    # Logit difference.
  d = delta - jnp.sum(p * delta, axis=-1, keepdims=True)               # Center under p.
  s = d - jnp.log1p(jnp.sum(p * jnp.expm1(d), axis=-1, keepdims=True)) # log(q / p).
  u = jnp.where(jnp.abs(s) < 1, p * jnp.expm1(s), q - p)               # Difference q - p.
  g = u - q * jnp.sum(u, axis=-1, keepdims=True)                       # Ensure sum(g) = 0.
  sg = jax.lax.stop_gradient
  loss = -jnp.sum(p * sg(log_q), axis=-1)                              # Forward loss.
  return loss + jnp.sum(sg(g) * (z_S - sg(z_S)), axis=-1)              # Custom gradient g.
\end{lstlisting}

As an example, let $\zstudent$ be a small \fpthirtytwo{} perturbation of
$\zteacher$.
Standard soft-target \crossentropytxt{} shrinks the second gradient
entry by nearly $3\times$, flips the sign of the third, and fails to sum to
zero, while our implementation matches the exact \fpsixtyfour{} gradient:

\begin{lstlisting}
z_T = jnp.array([2.0, 1.0, 0.0], dtype=jnp.float32)
z_S = z_T + jnp.array([-1e-7, 1e-7, 0.0], dtype=jnp.float32)
std_soft_ce = lambda z_S, z_T: -jnp.sum(jax.nn.softmax(z_T) * jax.nn.log_softmax(z_S))
jax.grad(std_soft_ce)(z_S, z_T)               # -> [-5.960e-8, 1.452e-8, -1.123e-9]
jax.grad(soft_target_cross_entropy)(z_S, z_T) # -> [-4.596e-8, 4.144e-8,  4.513e-9]
# Exact (float64) gradient:                        [-4.596e-8, 4.144e-8,  4.513e-9]
\end{lstlisting}

\section{The Slingshot Mechanism}
\label{sec:slingshot}

In grokking \citep{power2022grokking}, a network quickly memorizes the training
set ($\predprob_\targetidx \to 1$) and generalizes much later.
While studying grokking on modular addition with Adam
\citep{kingma2015adam,loshchilov2019decoupled}, \citet{thilak2022slingshot}
observed the \emph{slingshot mechanism}, where training after memorization
alternates between flat loss plateaus near zero and sudden loss spikes that
coincide with jumps in test accuracy.

\begin{figure}[t!]
    \centering
    \includegraphics[width=0.46\textwidth]{%
      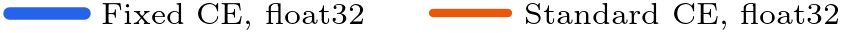%
    }\\[2pt]
    \begin{subfigure}[t]{0.3451\textwidth}
        \centering
        \includegraphics[width=\textwidth]{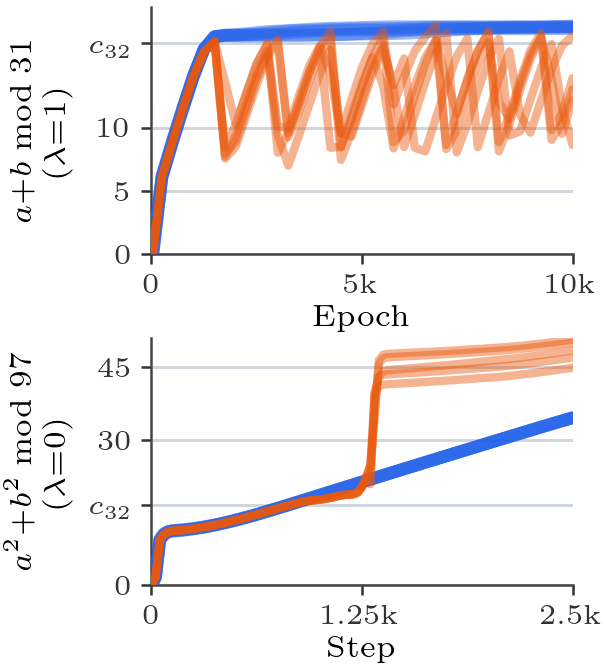}
        \captionsetup{margin={0.1065\textwidth,0pt}}
        \caption{Mean logit margin $\margin$}
        \label{fig:fig_slingshot_a}
    \end{subfigure}\hfill
    \begin{subfigure}[t]{0.3183\textwidth}
        \centering
        \includegraphics[width=\textwidth]{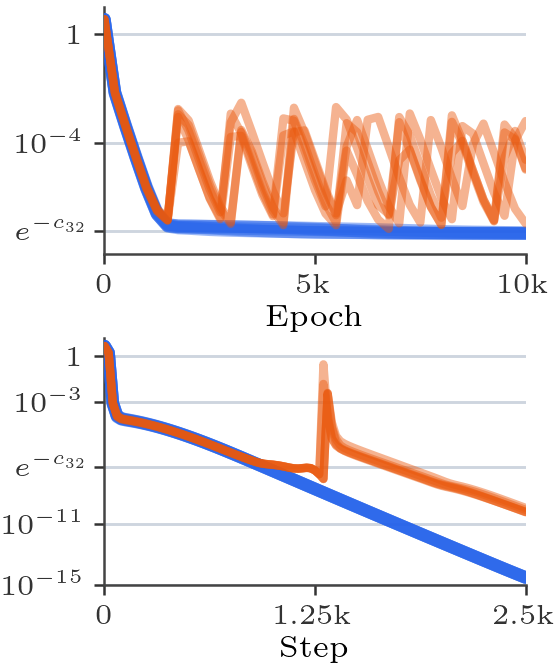}
        \captionsetup{margin={0.0797\textwidth,0pt}}
        \caption{Training loss}
        \label{fig:fig_slingshot_b}
    \end{subfigure}\hfill
    \begin{subfigure}[t]{0.3166\textwidth}
        \centering
        \includegraphics[width=\textwidth]{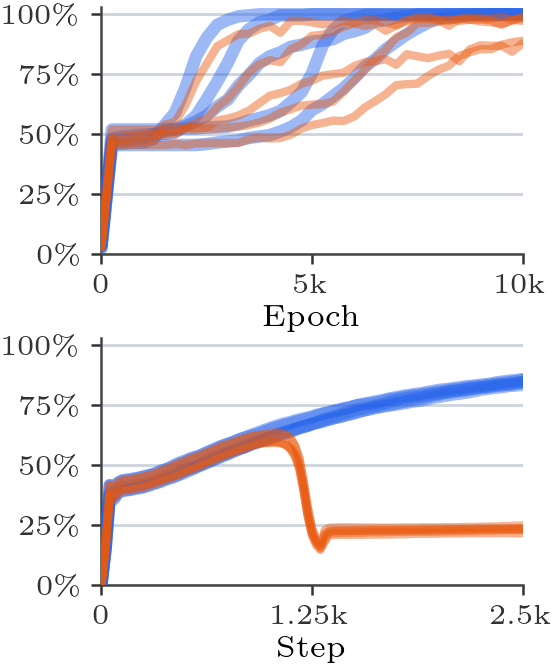}
        \captionsetup{margin={0.0780\textwidth,0pt}}
        \caption{Test accuracy}
        \label{fig:fig_slingshot_c}
    \end{subfigure}
    \vspace{-4pt}
    \caption{%
      The slingshot mechanism \citep{thilak2022slingshot} with and without
      weight decay (5 random seeds, \fpthirtytwo{}).
      \textbf{Top row ($a + b \bmod 31$, $\wdparam = 1$, $\adameps = 10^{-8}$,
      \citealt{varma2023explaining}):} Standard cross-entropy hits the
      \fpthirtytwo{} cliff ($\margin = \cliffthirtytwo \approx 16.6$, where loss
      rounds to $0$) and repeatedly drops and rebounds, slowing and degrading
      grokking; fixed cross-entropy crosses $\cliffthirtytwo$ smoothly, groks
      faster on all 5 seeds, and saturates just above $\cliffthirtytwo$ near
      $\log(\sfrac{1}{\adameps}) \approx 18.4$ (Section~\ref{sec:equilibria}).
      \textbf{Bottom row ($a^2 + b^2 \bmod 97$, $\wdparam = 0$, $\adameps =
      10^{-16}$, \citealt{doshi2024grokking}):} Without weight decay, hitting
      $\cliffthirtytwo$ around step $1250$ triggers a spike that drops standard
      cross-entropy test accuracy from $61\%$ to $23\%$, whereas fixed
      cross-entropy continues smoothly to $85\%$.
    }
    \label{fig:fig_slingshot}
    \vspace{-8pt}
\end{figure}

These slingshot spikes are a direct consequence of Softmax Collapse
(Section~\ref{sec:bug1}, \citealt{prieto2025grokking,liu2026grokking}).
Recall that AdamW updates parameters $\vparams$ by subtracting
$\lrparam\big(\vadamfirst / (\sqrt{\vadamsecond} + \adameps) +
\wdparam\vparams\big)$, where $\vadamfirst$ and $\vadamsecond$ are
bias-corrected moving averages of the parameter gradient $\vgrad =
\nabla_{\vparams}\loss(\vparams)$ and its square (with decay rates $\betaone,
\betatwo$).
After memorization, the logit margins $\margin$ grow as the network pushes its
target probabilities toward $1$.
Once an example's margin exceeds the precision cliff $\cliffd$, standard
cross-entropy rounds its target gradient $\fgrad_\targetidx$ to zero while its
non-target gradients $\fgrad_j > 0$ ($j \neq \targetidx$) remain positive.
When all examples in a batch sit above $\cliffd$, the parameter gradients nearly
vanish and Adam's second moment $\adamsecond_i$ decays exponentially toward zero
while weight decay slowly shrinks the parameters.
Eventually, weight decay or the residual non-target updates pull a few examples
back below $\cliffd$, where their target gradients $\fgrad_\targetidx \approx
-e^{-\margin}$ abruptly turn back on.
Because $\sqrt{\adamsecond_i}$ decayed while those examples were above
$\cliffd$, Adam divides the restored gradients by a very small denominator
$\sqrt{\adamsecond_i} + \adameps$, producing a large parameter step that spikes
the training loss, resets $\adamsecond_i$, and repeats the cycle.
With our fixed cross-entropy, the target-logit gradient $\grad_\targetidx =
-e^{-\margin}/(1 + e^{-\margin})$ does not round to zero at $\cliffd$, so
$\adamfirst_i$ and $\sqrt{\adamsecond_i}$ decay smoothly together and slingshot
spikes disappear (Figure~\ref{fig:fig_slingshot}).

Whether slingshot spikes appear to help or hurt generalization depends on how
the network's weight norm is controlled (Figure~\ref{fig:fig_slingshot}).
When a network uses LayerNorm before the output layer and no weight decay
\citep{thilak2022slingshot,prieto2025grokking}, growing the pre-LayerNorm
weights does not change the logits, but it does shrink the effective step size
$\lrparam / \|\vparams\|_2$; each slingshot spike inflates $\|\vparams\|_2$ and
acts as an accidental learning-rate decay.
When weight norms are already controlled by weight decay
\citep{varma2023explaining}, slingshot spikes at $\cliffthirtytwo$ only slow
down and hurt grokking ($94 \pm 5\%$ vs.\ $100 \pm 0\%$ test accuracy on $a + b
\bmod 31$; Figure~\ref{fig:fig_slingshot}, top).
And when neither LayerNorm nor weight decay is used \citep{doshi2024grokking}, a
single slingshot spike at $\cliffthirtytwo$ is destructive: on $a^2 + b^2 \bmod
97$ ($\wdparam = 0$, $\adameps = 10^{-16}$), standard cross-entropy tracks fixed
cross-entropy up to $61\%$ test accuracy until $\margin$ hits $\cliffthirtytwo$
around step $1250$, where a spike drops test accuracy to $23\%$ (and from $60\%$
to $21\%$ on $a^3 + b \bmod 97$), while fixed cross-entropy continues smoothly
to $85\%$ (Figure~\ref{fig:fig_slingshot}, bottom).

\section{Margin Equilibria}
\label{sec:equilibria}

\begin{figure}[h!]
    \vspace{-6pt}
    \centering
    \begin{minipage}[c]{0.495\textwidth}
        \centering
        \includegraphics[width=0.88\textwidth]{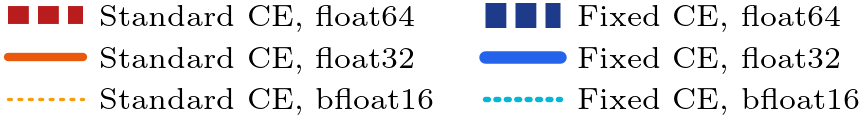}\\[4pt]
        \caption{%
          Normalized Adam step
          $|\grad_\targetidx| / (|\grad_\targetidx| + \adameps)$ vs.\ logit
          margin $\margin$ at $\adameps = 10^{-8}$.
          With our fixed cross-entropy, the step stays near $1$ and decays
          smoothly around $\log(\sfrac{1}{\adameps}) \approx 18.4$, whereas
          standard cross-entropy zeroes out $\fgrad_\targetidx$ at
          $\cliffsixteen \approx 5.5$ and $\cliffthirtytwo \approx 16.6$.
        }
        \label{fig:fig1b}
    \end{minipage}\hfill
    \begin{minipage}[c]{0.475\textwidth}
        \centering
        \includegraphics[width=\textwidth]{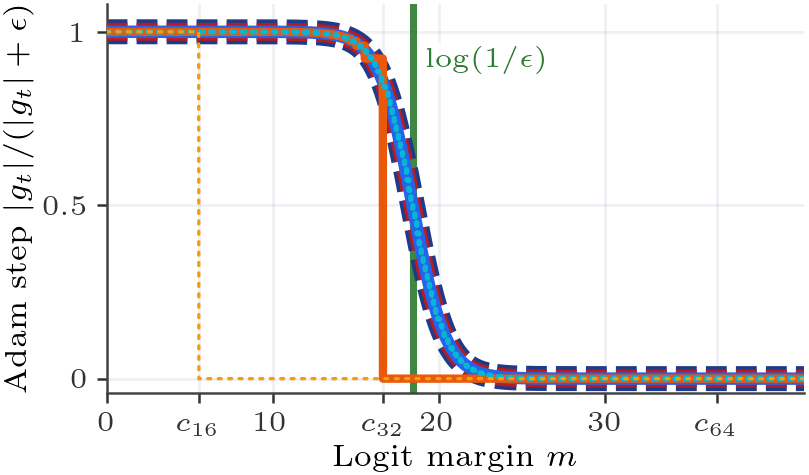}
    \end{minipage}
\end{figure}

In Figure~\ref{fig:fig_slingshot}a, our fixed cross-entropy eliminates the
precision cliff at $\cliffthirtytwo \approx 16.6$, yet with weight decay
($\wdparam = 1$, top) its margin still saturates just above $\cliffthirtytwo$
near $18$, whereas without weight decay ($\wdparam = 0$, bottom) its margin
continues to climb.
This plateau is the equilibrium where weight decay balances AdamW's normalized
gradient step.
At a stationary point where the parameters stop changing, the moving averages
converge to $\adamfirst_i = \grad_i$ and $\adamsecond_i = \grad_i^2$ (so
$\sqrt{\adamsecond_i} = |\grad_i|$), and setting the update to zero gives:
\begin{equation}
    \frac{\adamfirst_i}{\sqrt{\adamsecond_i} + \adameps} + \wdparam\params_i = 0
    \quad\Longrightarrow\quad
    \frac{\grad_i}{|\grad_i| + \adameps} + \wdparam\params_i = 0
    \quad \forall i\,.
    \label{eq:fp}
\end{equation}
Here $\lrparam$, $\betaone$, and $\betatwo$ cancel out, so RMSProp
\citep{tieleman2012rmsprop} and LaProp \citep{ziyin2020laprop} share the same
stationary point.
After memorization, the gradients shrink exponentially with the margin
($|\grad_i| \propto e^{-\margin}$).
If $\adameps = 0$, the normalized gradient $\grad_i / |\grad_i| =
\mathrm{sign}(\grad_i)$ has a magnitude of $1$ no matter how small $|\grad_i|$
becomes, which causes the margin to grow until the gradients underflow.
When $\adameps > 0$, however, the normalized gradient has magnitude $\approx 1$
while $|\grad_i| \gg \adameps$, and shrinks as $\grad_i / \adameps \propto
e^{-\margin}$ once $|\grad_i| \ll \adameps$ (Figure~\ref{fig:fig1b}).
Balancing this shrinking step against weight decay $\wdparam\params_i$ requires
$|\grad_i| \approx \adameps\wdparam|\params_i|$
\citep{wortsman2023small,molybog2023theory,everett2024scaling,%
kosson2024rotational}, which acts like an $L_2$ regularizer with weight
$\adameps\wdparam$ \citep{rosset2004margin,soudry2018implicit}.
Since $|\grad_i| \propto e^{-\margin}$, the margin stops growing at:
\begin{equation}
    \eqmargin(\adameps) \approx \log(\sfrac{1}{\adameps})\,.
    \label{eq:margin_law}
\end{equation}

\begin{figure}[h!]
    \vspace{-8pt}
    \centering
    \begin{minipage}[c]{0.605\textwidth}
        \centering
        \begin{subfigure}[t]{0.5287\textwidth}
            \centering
            \includegraphics[width=\textwidth]{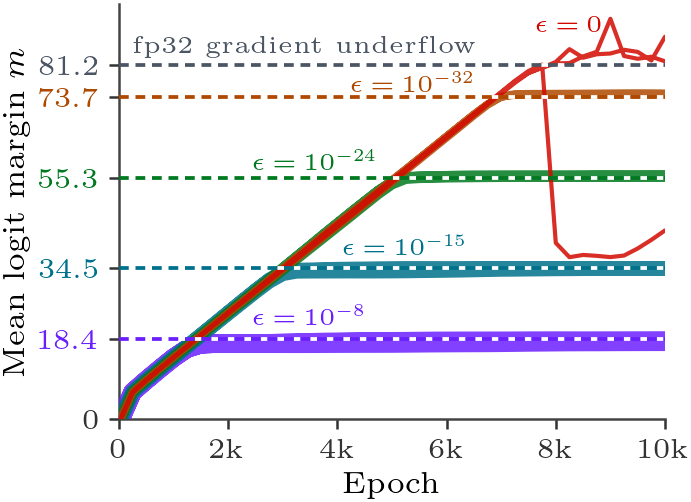}
            \vspace{-4pt}
            \captionsetup{margin={0.0739\textwidth,0pt}}
            \caption{Weight decay ($\wdparam = 1$)}
            \label{fig:fig6a}
        \end{subfigure}\hfill
        \begin{subfigure}[t]{0.4463\textwidth}
            \centering
            \includegraphics[width=\textwidth]{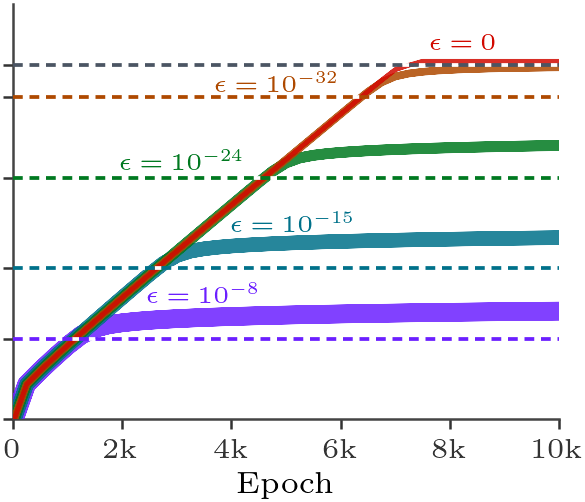}
            \vspace{-4pt}
            \caption{No weight decay ($\wdparam = 0$)}
            \label{fig:fig6b}
        \end{subfigure}
    \end{minipage}\hfill
    \begin{minipage}[c]{0.37\textwidth}
        \caption{%
          Logit margins with our fixed cross-entropy and LaProp on $a + b \bmod
          31$ in \fpthirtytwo{} ($\betaone = 0.9$, $\betatwo = 0.98$; 3 seeds
          per setting as solid lines, analytical predictions
          $\eqmargin(\adameps) = \log(\sfrac{1}{\adameps})$ as dashed lines).
          \textbf{(a)} With weight decay ($\wdparam = 1$), margins level off at
          $\log(\sfrac{1}{\adameps})$.
          \textbf{(b)} Without weight decay ($\wdparam = 0$), margins overshoot
          $\log(\sfrac{1}{\adameps})$.
        }
        \label{fig:fig6}
    \end{minipage}
    \vspace{-8pt}
\end{figure}

Equation~\ref{eq:margin_law} explains why the default $\adameps = 10^{-8}$
causes slingshot spikes in \fpthirtytwo{}: its equilibrium
$\log(\sfrac{1}{\adameps}) \approx 18.4$ sits just above $\cliffthirtytwo
\approx 16.6$ (Figure~\ref{fig:fig1b}), so standard cross-entropy hits the cliff
before equilibrium (whereas for $\adameps \ge 10^{-7}$,
$\log(\sfrac{1}{\adameps}) \le 16.1 < \cliffthirtytwo$ and standard
cross-entropy settles without spiking).
To test Equation~\ref{eq:margin_law} across many orders of magnitude down to
$\adameps = 0$ in \fpthirtytwo{} (Figure~\ref{fig:fig6}), we train with our
fixed cross-entropy and LaProp on $a + b \bmod 31$, updating
$\sqrt{\vadamsecond}$ directly without squaring $\grad_i$ (which underflows in
\fpthirtytwo{} once $|\grad_i| < 2^{-63}$, or $\margin \approx 38$).
With weight decay ($\wdparam = 1$, Figure~\ref{fig:fig6a}), the margin for each
$\adameps > 0$ climbs along the $\adameps = 0$ trajectory and levels off at
$\eqmargin(\adameps) = \log(\sfrac{1}{\adameps})$, while $\adameps = 0$ grows
until the \fpthirtytwo{} gradients underflow at $\margin \approx 81.2$.
Without weight decay ($\wdparam = 0$, Figure~\ref{fig:fig6b}), nothing balances
the $\grad_i / \adameps$ step, so margins keep rising past
$\log(\sfrac{1}{\adameps})$.

\section{Grokking Across Precisions and Benchmarks}
\label{sec:experiments}

\begin{figure}[h!]
    \centering
    \includegraphics[width=0.68\textwidth]{paper_fig2_legend.png}\\[2pt]
    \begin{subfigure}[t]{0.3315\textwidth}
        \centering
        \includegraphics[width=\textwidth]{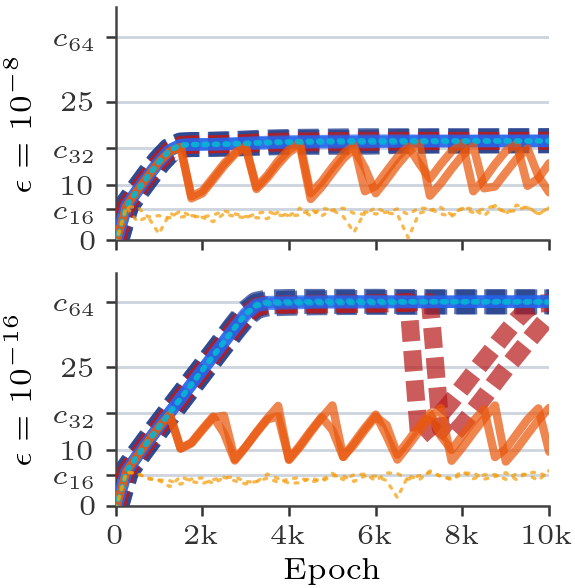}
        \captionsetup{margin={0.0824\textwidth,0pt}}
        \caption{Mean logit margin $\margin$}
        \label{fig:fig2a}
    \end{subfigure}\hfill
    \begin{subfigure}[t]{0.3251\textwidth}
        \centering
        \includegraphics[width=\textwidth]{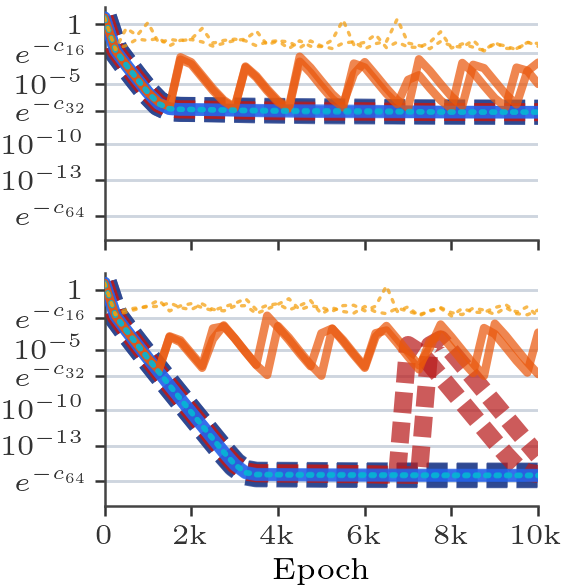}
        \captionsetup{margin={0.0698\textwidth,0pt}}
        \caption{Training loss}
        \label{fig:fig2b}
    \end{subfigure}\hfill
    \begin{subfigure}[t]{0.3234\textwidth}
        \centering
        \includegraphics[width=\textwidth]{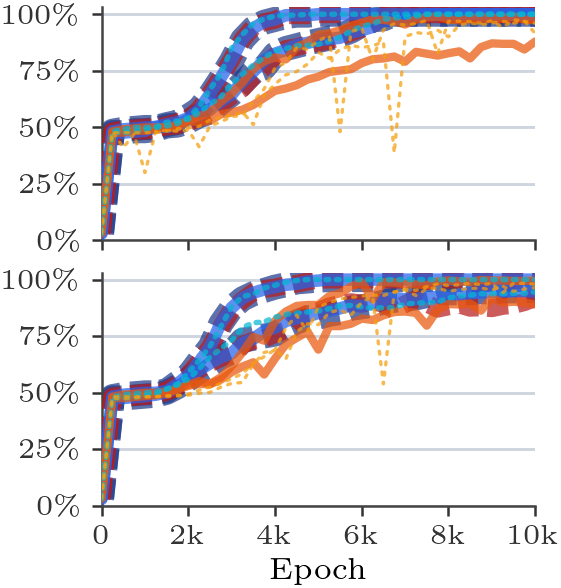}
        \captionsetup{margin={0.0703\textwidth,0pt}}
        \caption{Test accuracy}
        \label{fig:fig2c}
    \end{subfigure}
    \vspace{-4pt}
    \caption{%
      Standard cross-entropy fails at the precision cliff $\cliffd$; fixed
      cross-entropy removes the cliff across all precisions ($a + b \bmod 31$,
      $\wdparam = 1$; 2 of 5 seeds shown).
      Top: $\adameps = 10^{-8}$.
      Bottom: $\adameps = 10^{-16}$, where even \fpsixtyfour{} standard
      cross-entropy fails at $\cliffsixtyfour$ while fixed cross-entropy
      converges smoothly.
    }
    \label{fig:fig2}
    \vspace{-6pt}
\end{figure}

To see how the precision cliff behaves across floating-point formats, we follow
\citet{varma2023explaining} and \citet{lyle2025grokking} and train a 1-layer,
4-head, width-$64$ ReLU Transformer without LayerNorm on $a + b \bmod 31$
($50\%$ training split) for $10{,}000$ epochs of full-batch AdamW ($\lrparam =
10^{-3}$, $\betaone = 0.9$, $\betatwo = 0.98$, $\wdparam = 1$), varying the loss
(standard vs.\ fixed cross-entropy), logit precision at the loss (\bfsixteen{},
\fpthirtytwo{}, \fpsixtyfour{}), and $\adameps \in \{10^{-8}, 10^{-12},
10^{-16}\}$ (Figure~\ref{fig:fig2}, Table~\ref{tab:baselines}).
At $\adameps = 10^{-8}$ (Figure~\ref{fig:fig2}, top), standard cross-entropy
stalls at $\cliffsixteen \approx 5.5$ in \bfsixteen{} and repeatedly spikes at
$\cliffthirtytwo \approx 16.6$ in \fpthirtytwo{}, whereas fixed cross-entropy
crosses both cliffs smoothly to $\log(\sfrac{1}{\adameps}) \approx 18.4$ and
reaches $100\%$ test accuracy in all three precisions.
Reducing $\adameps$ to $10^{-16}$ (Figure~\ref{fig:fig2}, bottom) raises the
equilibrium to $\log(\sfrac{1}{\adameps}) \approx 36.8 > \cliffsixtyfour$,
breaking even \fpsixtyfour{} standard cross-entropy at $\cliffsixtyfour \approx
36.7$ while fixed cross-entropy converges smoothly across all precisions.

Table~\ref{tab:baselines} summarizes these results and compares against two
recent methods that modify the gradients or loss to mitigate Softmax Collapse.
\citet{liu2026grokking} subtract the mean logit gradient from each entry so the
gradients sum to zero (``Zero-sum Proj.''), which eliminates slingshot spikes.
However, once $\margin > \cliffthirtytwo$ and $\fgrad_\targetidx = 0$,
subtracting the mean only gives the target logit $-\sfrac{1}{\numclasses}$ of
the total non-target gradient, stalling $\margin$ near $18$ and dropping test
accuracy below $50\%$ for $\adameps \le 10^{-12}$.
\citet{prieto2025grokking} propose ``StableMax,'' which replaces $\exp(x)$ in
\softmaxtxt{} with a piecewise rational function ($x + 1$ for $x \ge 0$ and
$1/(1 - x)$ for $x < 0$).
Because this function grows linearly instead of exponentially, logit margins
blow up to ${\sim}10^4$ and test accuracy lags our fixed cross-entropy across
all $\adameps$.

\begin{figure}[h!]
    \vspace{-8pt}
    \centering
    \includegraphics[width=0.62\textwidth]{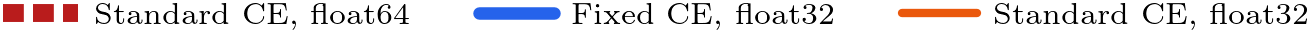}\\[2pt]
    \begin{subfigure}[t]{0.2969\textwidth}
        \centering
        \includegraphics[width=\textwidth]{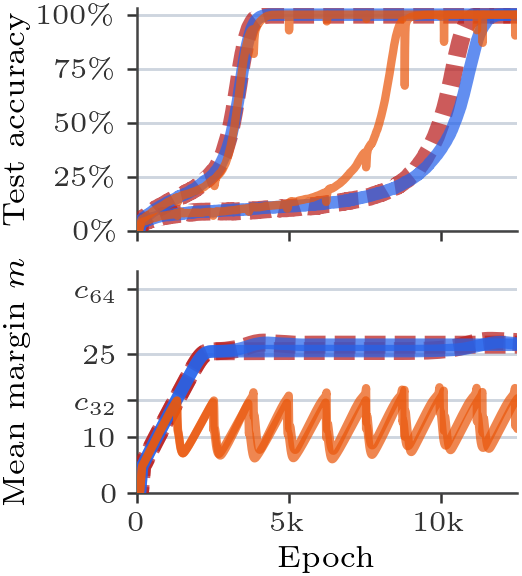}
        \captionsetup{margin={0.0705\textwidth,0pt}}
        \caption{$a{+}b \bmod 113$ \\ \citet{nanda2023progress}}
        \label{fig:fig3a}
    \end{subfigure}\hfill
    \begin{subfigure}[t]{0.2264\textwidth}
        \centering
        \includegraphics[width=\textwidth]{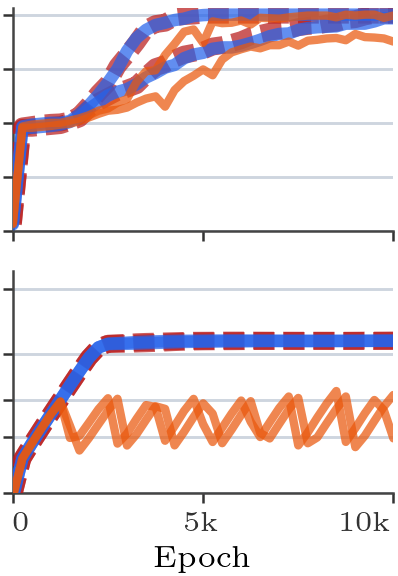}
        \caption{$a{+}b \bmod 31$ \\ \citet{varma2023explaining}}
        \label{fig:fig3b}
    \end{subfigure}\hfill
    \begin{subfigure}[t]{0.2264\textwidth}
        \centering
        \includegraphics[width=\textwidth]{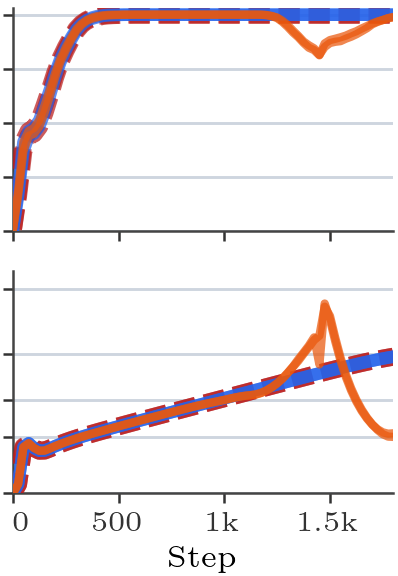}
        \caption{$a^2{+}b^2 \bmod 97$ \\ \citet{doshi2024grokking}}
        \label{fig:fig3c}
    \end{subfigure}\hfill
    \begin{subfigure}[t]{0.2264\textwidth}
        \centering
        \includegraphics[width=\textwidth]{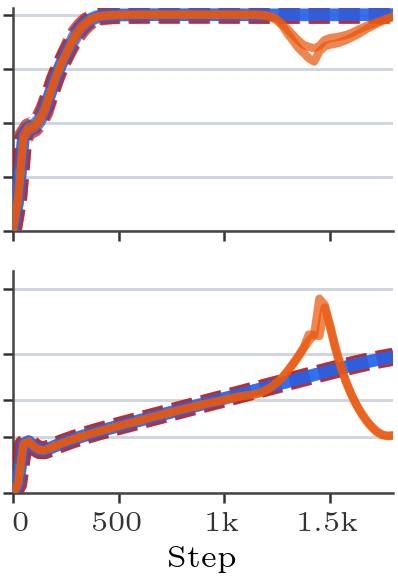}
        \caption{$a^3{+}b \bmod 97$ \\ \citet{doshi2024grokking}}
        \label{fig:fig3d}
    \end{subfigure}
    \vspace{-4pt}
    \caption{%
      Results across four grokking tasks at $\adameps = 10^{-12}$ (2 of 3--5
      seeds shown): 1-layer Transformers (a, b) and 2-layer MLPs (c, d).
      Fixed cross-entropy in \fpthirtytwo{} (blue) matches \fpsixtyfour{}
      standard cross-entropy (red), while standard cross-entropy in
      \fpthirtytwo{} (orange) breaks at $\cliffthirtytwo$.
    }
    \label{fig:fig3}
    \vspace{6pt}
\end{figure}

We observe the same behavior across four grokking benchmarks at $\adameps =
10^{-12}$ (Figure~\ref{fig:fig3}): 1-layer Transformers on $a + b \bmod 113$
\citep{nanda2023progress} and $a + b \bmod 31$ \citep{varma2023explaining} with
$\wdparam = 1$, and 2-layer MLPs on $a^2 + b^2 \bmod 97$ and $a^3 + b \bmod 97$
\citep{doshi2024grokking} with $\wdparam = 0.2$.
On all four tasks, \fpthirtytwo{} standard cross-entropy spikes once it hits
$\cliffthirtytwo$, whereas \fpthirtytwo{} fixed cross-entropy matches
\fpsixtyfour{} standard cross-entropy and reaches $100\%$ test accuracy.

\begin{table}[h!]
\vspace{-4pt}
\centering
\footnotesize
\setlength{\tabcolsep}{3.6pt}%
\setlength{\aboverulesep}{1.0pt}%
\setlength{\belowrulesep}{1.4pt}%
\begin{tabular}{l@{\;\;}l *{3}{r@{\;}c@{\;}l} ccc}
 & & \multicolumn{9}{c}{Test Accuracy (\%)} & \multicolumn{3}{c}{Final Margin $\margin$} \\
\cmidrule(lr){3-11} \cmidrule(lr){12-14}
 & Variant\hfill\rlap{\hspace{5pt}$\adameps =$} & \multicolumn{3}{c}{$10^{-8}$} & \multicolumn{3}{c}{$10^{-12}$} & \multicolumn{3}{c}{$10^{-16}$} & $10^{-8}$ & $10^{-12}$ & $10^{-16}$ \\
\midrule
\fpthirtytwo{} & Fixed CE & 100.0 & $\pm$ & 0.0 & 99.6 & $\pm$ & 0.7 & 99.0 & $\pm$ & 2.3 & 17.9 & 27.4 & 36.8 \\
\fpthirtytwo{} & Standard CE & 94.3 & $\pm$ & 5.4 & 96.4 & $\pm$ & 5.1 & 96.5 & $\pm$ & 4.3 & 13.0 & 15.1 & 12.7 \\
\midrule
\bfsixteen{} & Fixed CE & 100.0 & $\pm$ & 0.1 & 99.8 & $\pm$ & 0.2 & 98.9 & $\pm$ & 2.4 & 17.9 & 27.4 & 36.8 \\
\bfsixteen{} & Standard CE & 93.3 & $\pm$ & 7.4 & 94.9 & $\pm$ & 4.9 & 94.2 & $\pm$ & 8.3 & 5.1 & 5.6 & 5.8 \\
\midrule
\fpsixtyfour{} & Fixed CE & 100.0 & $\pm$ & 0.0 & 99.5 & $\pm$ & 1.1 & 99.0 & $\pm$ & 2.0 & 17.9 & 27.4 & 36.8 \\
\fpsixtyfour{} & Standard CE & 100.0 & $\pm$ & 0.0 & 100.0 & $\pm$ & 0.1 & 97.2 & $\pm$ & 3.2 & 17.9 & 27.4 & 30.1 \\
\midrule
\fpthirtytwo{} & Zero-sum Proj. \citep{liu2026grokking} & 97.9 & $\pm$ & 3.2 & 48.9 & $\pm$ & 2.6 & 49.3 & $\pm$ & 2.5 & 17.0 & 18.9 & 18.9 \\
\fpthirtytwo{} & StableMax \citep{prieto2025grokking} & 97.3 & $\pm$ & 4.7 & 95.6 & $\pm$ & 2.0 & 97.3 & $\pm$ & 4.3 & 3200 & 5661 & 6313 \\
\fpsixtyfour{} & StableMax \citep{prieto2025grokking} & 97.4 & $\pm$ & 4.2 & 96.7 & $\pm$ & 5.3 & 95.6 & $\pm$ & 2.7 & 3170 & 21252 & 2322 \\
\end{tabular}
\caption{%
  Comparison of loss functions on modular addition ($a + b \bmod 31$, $\wdparam
  = 1$, $3$--$5$ seeds) across $\adameps \in \{10^{-8}, 10^{-12}, 10^{-16}\}$.
  Fixed CE reaches the predicted equilibrium margin ($\log(\sfrac{1}{\adameps})
  \approx 18.4, 27.6, 36.8$) and ${\ge}98.9\%$ test accuracy across all
  precisions, whereas Standard CE stalls once $\eqmargin(\adameps)$ exceeds the
  cliff $\cliffd$ ($\cliffsixteen \approx 5.5$ in \bfsixteen{},
  $\cliffthirtytwo \approx 16.6$ in \fpthirtytwo{}, and $\cliffsixtyfour
  \approx 36.7$ in \fpsixtyfour{}).
}
\label{tab:baselines}
\vspace{-8pt}
\end{table}

\vspace{-4pt}
\section{Soft-Target Distillation}
\label{sec:distillation}

\begin{figure}[h!]
    \centering
    \includegraphics[width=0.70\textwidth]{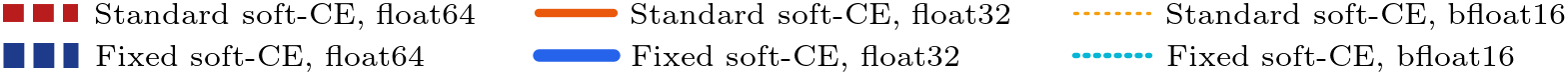}\\[2pt]
    \begin{subfigure}[t]{0.4843\textwidth}
        \centering
        \includegraphics[width=\textwidth]{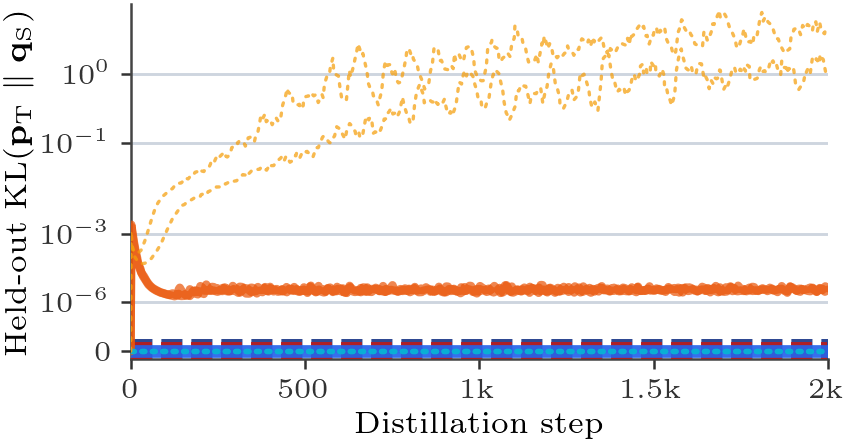}
        \captionsetup{margin={0.0651\textwidth,0pt}}
        \caption{Held-out $\mathrm{KL}(\pteacher \parallel \qstudent)$}
        \label{fig:fig4b}
    \end{subfigure}\hfill
    \begin{subfigure}[t]{0.4877\textwidth}
        \centering
        \includegraphics[width=\textwidth]{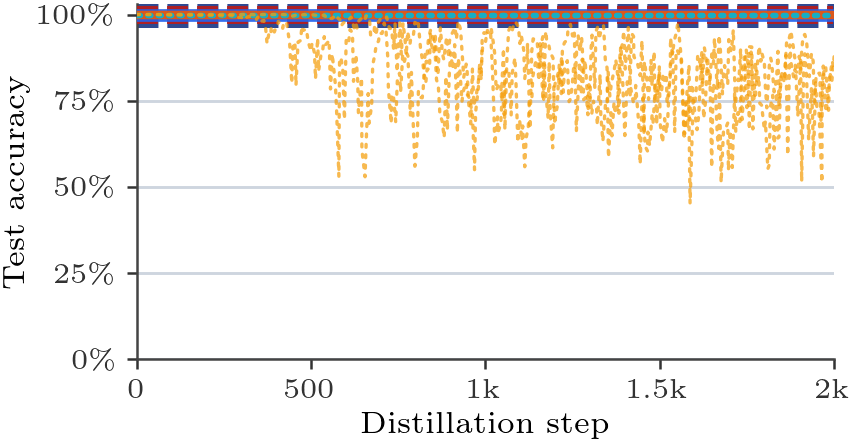}
        \captionsetup{margin={0.0685\textwidth,0pt}}
        \caption{Test accuracy}
        \label{fig:fig4c}
    \end{subfigure}
    \vspace{-4pt}
    \caption{%
      Soft-target distillation when the student is initialized with the
      teacher's weights ($\paramsstudent^{(0)} = \paramsteacher$, $\temp = 4$,
      $\adameps = 10^{-8}$, 2 of 3 seeds shown).
      Fixed soft-target cross-entropy stays at $\mathrm{KL} = 0$ and $100\%$
      accuracy across all precisions; standard soft-target cross-entropy
      amplifies rounding error into parameter drift in \bfsixteen{} and an
      initial $\mathrm{KL}$ spike and ${\sim}10^{-6}$ floor in \fpthirtytwo{}.
    }
    \label{fig:fig4}
\end{figure}

To see how Softmax Cancellation (Section~\ref{sec:bug2}) affects optimization,
we initialize a 2-layer ReLU MLP student with the \fpthirtytwo{} weights of a
teacher trained to $100\%$ accuracy on $a + b \bmod 53$ ($\paramsstudent^{(0)} =
\paramsteacher$), and run $2{,}000$ full-batch steps of Adam ($\lrparam =
2\times 10^{-3}$, $\betaone = 0.9$, $\betatwo = 0.98$, $\wdparam = 0$,
$\adameps = 10^{-8}$) on the $50\%$ training split at temperature $\temp = 4$
\citep{hinton2015distilling} (which keeps $\margin/\temp < \cliffsixteen$ so
$\vtargetprob$ does not round to one-hot), computing the forward pass and loss
in \bfsixteen{}, \fpthirtytwo{}, or \fpsixtyfour{}.
Because $\zstudent = \zteacher$ initially, the true gradient is zero and the
student should not move.
However, standard soft-target cross-entropy has a non-zero residual gradient at
$\zstudent = \zteacher$ due to the mismatch between its forward \softmaxtxt{}
and backward \logsoftmaxtxt{} (Figure~\ref{fig:fig4a}).
Because Adam divides by $\sqrt{\vadamsecond} + \adameps$, it amplifies even tiny
residual gradients (Figure~\ref{fig:fig4}): in \bfsixteen{} (${\sim}10^{-3}$) it
drives the student away from the teacher and degrades test accuracy, in
\fpthirtytwo{} (${\sim}10^{-8}$) it causes an initial $\mathrm{KL}$ spike and a
${\sim}10^{-6}$ floor, and in \fpsixtyfour{} (${\sim}10^{-17}$) the update is
too small to change the \fpthirtytwo{} weights.
Our fixed soft-target cross-entropy returns $\vgrad = \vzero$ at $\zstudent =
\zteacher$, so the student stays at $\mathrm{KL} = 0$ and $100\%$ accuracy
across all precisions.

\section{Conclusion}
\label{sec:recommendations}

We have presented simple fixes for two numerical bugs in standard
implementations of \softmaxtxt{}, \logsoftmaxtxt{}, and \crossentropytxt{}:
\emph{Softmax Collapse}, which zeroes out the dominant logit's gradient at high
confidence while leaving non-dominant gradients non-zero, and \emph{Softmax
Cancellation}, which loses precision when student logits approach teacher logits
in soft-target distillation.
Rewriting these functions in log space using $\softplus$, $\expmone$, and
$\logonep$ restores accurate, zero-sum gradients across \bfsixteen{},
\fpthirtytwo{}, and \fpsixtyfour{} without internally casting to higher
precision.
Fixing these bugs eliminates slingshot loss spikes during grokking, prevents
student networks from drifting away from the teacher during distillation, and
reveals that logit margins under AdamW and weight decay naturally converge to
$\log(\sfrac{1}{\adameps})$.
Though the extreme margins and near-zero student--teacher gaps probed by our
experiments are uncommon in standard large-scale training, our fixed functions
are easy drop-in replacements that can be implemented without meaningful
slowdown, and may prove useful whenever models make high-confidence predictions
or distill from close teachers.

\textbf{Acknowledgments:}
We thank Jean-Baptiste Alayrac, Forrester Cole, Roy Frostig, Mark Hamilton,
Muhammad Maaz, Miki Rubinstein, Keith Rush, and Brandon Zhao for their helpful
advice.
\vfill\newpage

\textbf{Errata:}
In an earlier version of this paper, the code in Section~\ref{sec:bug1}
produced $\mathrm{NaN}$ gradients when all but one logit was effectively
$-\infty$, as happens with masking in causal attention or with extreme logits.
This bug was not present in our internal codebase, which produced all results
in this paper.
It appears to have been introduced by an over-eager agent that simplified our
code for this paper, likely in an attempt to match
Equations~\ref{eq:logsoftmax_fix} and~\ref{eq:hard_fix} (with $s = \sum_{j
\neq \imax} e^{\shiftedlogits_j}$, $\softplus(-\margin)$ can be written as
$\logonep(s)$ or equivalently as $\softplus(\log s)$, but the latter makes
autodiff differentiate through $\log 0$ when $s = 0$).
We thank \citet{maaz2026property} for identifying this issue.

\bibliography{references}
\bibliographystyle{iclr2026_conference}

\end{document}